\documentclass[journal]{IEEEtran}
\ifCLASSINFOpdf
\else
\fi
\usepackage{graphicx}
\usepackage{epstopdf}
\usepackage{subfigure}
\usepackage{amssymb,amsmath}
\usepackage{amsfonts}
\usepackage[ruled]{algorithm2e}
\usepackage{algpseudocode}
\usepackage{multirow}
\usepackage{multicol}
\usepackage[flushleft]{threeparttable}
\usepackage{balance}

\begin{document}
%
\title{Transductive Zero-Shot Learning with Adaptive Structural Embedding}
%
%
%

\author{Yunlong~Yu,
        Zhong~Ji,~\IEEEmembership{Member,~IEEE,}
        Jichang~Guo,
        and Yanwei Pang, \IEEEmembership{Senior Member,~IEEE}
}

\maketitle

\begin{abstract}
    Zero-shot learning (ZSL) endows the computer vision system with the inferential capability to recognize instances of a new category that has never seen before. Two fundamental challenges in it are visual-semantic embedding and domain adaptation in cross-modality learning and unseen class prediction steps, respectively. To address both challenges, this paper presents two corresponding methods named Adaptive STructural Embedding (ASTE) and Self-PAsed Selective Strategy (SPASS), respectively. Specifically, ASTE formulates the visual-semantic interactions in a latent structural SVM framework to adaptively adjust the slack variables to embody the different reliableness among training instances. In this way, the reliable instances are imposed with small punishments, wheras the less reliable instances are imposed with more severe punishments. Thus, it ensures a more discriminative embedding. On the other hand, SPASS offers a framework to alleviate the domain shift problem in ZSL, which exploits the unseen data in an easy to hard fashion. Particularly, SPASS borrows the idea from self-paced learning by iteratively selecting the unseen instances from reliable to less reliable to gradually adapt the knowledge from the seen domain to the unseen domain. Subsequently, by combining SPASS and ASTE, we present a self-paced Transductive ASTE (TASTE) method to progressively reinforce the classification capacity. Extensive experiments on three benchmark datasets (i.e., AwA, CUB, and aPY) demonstrate the superiorities of ASTE and TASTE. Furthermore, we also propose a fast training (FT) strategy to improve the efficiency of most of existing ZSL methods. The FT strategy is surprisingly simple and general enough, which can speed up the training time of most existing methods by 4$\thicksim$300 times while holding the previous performance.
\end{abstract}

%
\IEEEpeerreviewmaketitle

\section{Introduction}
%
%
%
%
\IEEEPARstart{R}{ecently}, image classification has made tremendous improvements due to the prosperous progress of deep learning and the availability of large scale annotated databases \cite{copr16:Pang}, \cite{tnnls14:Shao}. However, in many applications, it is impractical to obtain adequate labeled object categories \cite{tnnls16:Yu, tip15:Ji}, \cite{tnnls15: Shao}. To tackle this limitation, zero-shot learning (ZSL) \cite{cvpr09:Lampert,is17:Ji,cvpr16:Fu,pami15:Fu,cvpr15:Fu,cvpr13:Akata,cvpr15:Akata,cvpr16:Xian,icml15:Romera-Paredes} is proposed to recognize the unseen categories that no labeled data are available for training, i.e., the categories in training and testing are disjoint. It is inspired by the human beings' inferential ability that can recognize a novel class without seeing its visual instances and has received increasing attention in recent years.

The emergence of semantic vectors enables the realization of ZSL. By representing the class labels with semantic vectors, a semantic space is built to associate the semantic relationships between the seen and unseen categories. 
Thus, the knowledge from the seen categories can be transferred to the unseen categories within this space. Two popular semantic vectors used in ZSL are attributes \cite{cvpr09:Lampert,icml15:Romera-Paredes} and word vectors \cite{cvpr15:Akata, nips13:Socher, nips13:Frome}. Particularly, attributes define a few properties of objects, such as shape, color and the presence or absence of a certain body part, which are manually defined \cite{cvpr09:Lampert} or discriminatively learned \cite{cvpr13:Yu, wacv15:Al-Halah}. Word vectors represent class names with vectors based on a distributed language representation technique, such as Word2Vec \cite{nips13:Mikolov} and Glove \cite{emnlp14:Pennington}.

 With the class semantic vectors, the cross-modality relationships between the image visual features and the semantic vectors can be exploited by a visual-semantic embedding either from  the visual space to the semantic space \cite{cvpr16:Fu, nips13:Socher}, or vice versa \cite{iccv15:Kodirov, arXiv16:Shojaee} or a shared common space \cite{cvpr15:Akata,aaai16:Guo}. Many effective methods have been proposed to build the visual-semantic embedding, including linear-based \cite{iccv15:Kodirov}, nonlinear-based \cite{cvpr15:Lei, nips13:Socher}, bilinear-based \cite{icml15:Romera-Paredes, cvpr16:Qiao, aaai16:Guo} and max margin-based \cite{cvpr15:Akata, aistats15:Li, nips13:Frome} approaches. Specifically, the max margin-based methods employ a ranking function to measure the compatibility scores between the images and the class semantic vectors, in which a compatibility matrix is derived by enforcing the correct label to be ranked higher than any of the other labels. However, in such models, the seen instances are typically treated without counting for their different reliableness during training, in which the structural information of the seen data may be undermined. To address this problem, we formulate the visual-semantic embedding in a latent structural SVM framework to adaptively adjust the slack variables to distinguish the training data, where the reliable instances are imposed with small punishments while the less reliable instances are imposed with more severe punishments. In this way, the structural information in the seen data are effectively exploited by assessing their reliability and discriminability.


After the embedding step, the label of an unseen instance can be determined by performing a nearest neighbor (NN) search to match its visual feature against the candidate unseen categories in the embedding space. However, since the seen and unseen categories are different and potentially unrelated, the embedding is biased when it is directly applied to the unseen data. This is the well-known projection domain shift problem in ZSL \cite{pami15:Fu,iccv15:Kodirov}. To alleviate this bias, many transductive ZSL approaches have been developed to aggregate the unseen data together with the seen data to learn a more general visual-semantic embedding to improve classification performance \cite{is17:Ji, cvpr15:Fu, aaai16:Guo, arXiv16:Shojaee}. However, such approaches mainly focus on exploiting the structural information in the unseen data and the potential label information is disregarded or underestimated. Actually, although the unseen data are unlabeled, we can predicted their potential labels with the knowledge learned from the unseen data. To this end, we propose to exploit the potential unseen label information in an easy to hard fashion, which includes two steps: (1) learning the visual-semantic embedding with the labeled seen data; (2) gradually refining the visual-semantic embedding with the seen data and unseen data in an iterative way. At each iteration, the unseen data is firstly predicted with the current visual-semantic embedding and the reliable unseen instances are selected as pseudo labeled data with the self-paced selective strategy, and then the pseudo labeled data are added into the labeled data set to refine the visual-semantic embedding. In this way, the knowledge is adapted progressively from the seen domain to the unseen domain. Meanwhile, the potential labeled information of unseen data is exploited in a confident way, thus the domain shift problem can be readily addressed. Figure 1 illustrates the proposed transductive framework.

\begin{figure*}[t]
    \begin{center}
   \includegraphics[height=5.8cm,width=17.0cm]{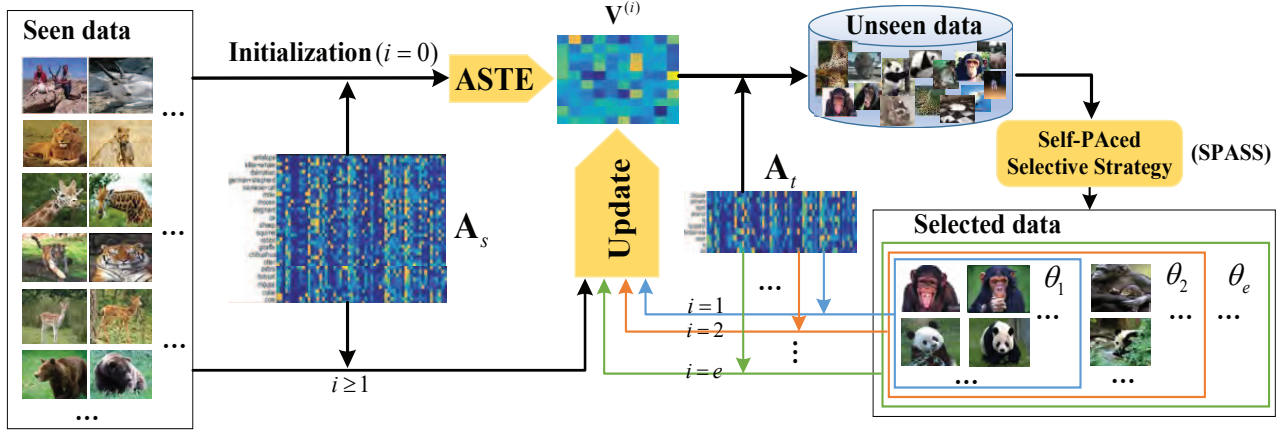}
    \end{center}
     \caption{Illustration of the proposed TASTE. $\mathbf{A}_s$ and $\mathbf{A}_t$ are class semantic matrixes (attributes or word vectors) for seen and unseen classes, respectively. $\mathbf{V}^{(t)}$ is the compatibility matrix at $t$-th iteration. $t=0$ means that only seen data are used for training, and $t\geq1$ means the model is refined iteratively. The initialized compatibility matrix $\mathbf{V}^{(0)}$ is trained with ASTE, of which only the seen data and the corresponding seen class semantic embeddings are available. Rather than predicting the unseen data in one pass, the proposed TASTE exploits the unseen data iteratively. At each iteration, the compatibility matrix is retrained with the labeled seen data and a set reliable unseen instances that are selected by SPASS. $i$ is the iterative number, which is controlled by the threshold $\theta_i$.}
    \label{fig:long}
    \label{fig:onecol}
  \end{figure*}

   The contributions of our work can be summarized as:
  \begin{enumerate}
    \item A novel Adaptive Structural Embedding (ASTE) method for visual-semantic embedding in ZSL is proposed, which formulates the embedding problem in a latent structural SVM framework to embody the different reliableness by adaptively adjusting the slack variables. Extensive experiments show it achieves comparable or better performance than state-of-the-art inductive methods.
    \item To alleviate the domain shift problem in ZSL, a Self-PAced Selective Strategy (SPASS) is presented, which iteratively selects a set of pseudo labeled instances from the unseen data to gradually refine the previously learned model. By combining this strategy and ASTE, we develop a Transductive ASTE (TASTE) method to progressively reinforce the classification capacity. TASTE outperforms most state-of-the-art transductive ZSL methods on three benchmark datasets: AwA, aPY, and CUB.
    \item Furthermore, to improve the training efficiency, we present a surprisingly simple but effective Fast Training (FT) strategy. It is based on an audacious idea that representing the visual features in each training category with their visual pattern. Extensive experiments demonstrate that it is a general strategy suitable to many existing ZSL methods, and can greatly increase the training speed of most existing methods by 4 to 300 times while hold their previous performances.
  \end{enumerate}

\section{Related work}
\subsection{Visual-semantic embedding for ZSL}
  Visual-semantic embedding is the key technique in ZSL, which bridges the relationships between the visual features and the semantic representations. It is learned from the seen data, while is applied to recognize the unseen data. Actually, it can be considered as a process of knowledge transfer or inference from the seen domain to the unseen domain.


  In recent years, several significant progresses have been made in the research of the visual-semantic embedding methods. One of the pioneering studies is \cite{cvpr09:Lampert}, where two fundamental visual-attributes embedding paradigms, i.e., Direct Attribute Prediction (DAP) and Indirect Attribute Prediction (IAP) are presented. Both of them use a probabilistic model to match the attribute prediction with the unseen categories. Specifically, IAP employs the attributes to connect the seen categories in a lower layer and the unseen categories in a higher layer, while DAP exploits the seen and unseen categories in one layer and connects them with attributes directly. In contrast, Socher \emph{et al.} \cite{nips13:Socher} and Frome \emph{et al.} \cite{nips13:Frome} are among the first researchers to establish the visual-word vector embedding. Specifically, Socher \emph{et al.} \cite{nips13:Socher} developed a two-layer neural networks to build a regression model for constructing interactional relationships between the visual space and the word vector space. It is a nonlinear method, which has not an explicit embedding matrix. Different from \cite{nips13:Socher}, DeViSE \cite{nips13:Frome} trains a linear mapping to link the image representation with the word vector using a combination of dot-product similarity and hinge rank loss, which is trained to produce a higher dot-product similarity between the visual feature and the word vector representation of the correct label than that of the visual feature and any other labels. Afterwards, several bilinear methods have also been proposed. For example, Structured Joint Embedding (SJE) \cite{cvpr15:Akata} relates the input embedding and output embedding through a compatibility function, and implements ZSL by finding the label corresponding to the highest joint compatibility score. Further, LatEm \cite{cvpr16:Xian} employs a bilinear compatibility model to learn a collection of maps as latent variables for the current image-class pair. The model is trained with a ranking based objective function that penalizes incorrect rankings of the true class for a given image.

  The studies similar to our ASTE method are \cite{icml15:Romera-Paredes} and \cite{aistats15:Li}. In \cite{icml15:Romera-Paredes}, an embarrassingly simple embedding method (ESZSL) is presented. It constructs a general framework to model the relationships between visual features, class attributes and class labels with a bilinear model, and the closed-form solution makes it efficient. As similar bilinear formulation is used as \cite{icml15:Romera-Paredes}, the proposed ASTE not only considers these relationships, but also captures the discriminantive inter-class information by penalizing the incorrect predictions with a max-margin model. In \cite{aistats15:Li}, the authors treat ZSL as a standard semi-supervised learning problem over the seen data and unsupervised clustering problem over the unseen data and integrate both parts in a latent max-margin multi-classification framework. Although using a similar max-margin multi-classification framework, the proposed ASTE learns a more discriminative embedding by distinguishing the different reliableness of the training data in a latent structural SVM framework. Furthermore, in our transductive framework, the potential label information of unseen data is gradually exploited by constructing the interaction between the seen and unseen data rather than exploiting the structural information of unseen data with an unsupervised clustering method.


\subsection{Domain shift in ZSL}
  ZSL can be viewed as a special case of transductive domain adaptation where the training and testing data have non-overlapping labels \cite{pami15:Fu, iccv15:Kodirov, eccv16:Xu}. Since the embedding function learned from the seen data is biased when directly applied to the disjoined unseen data, the domain shift problem occurs \cite{pami15:Fu,iccv15:Kodirov,icip15:Xu,eccv16:Xu}. It will decrease the classification performance on the unseen data. Different from the conventional domain shift problem \cite{iccv13:Fernando,cvpr09:Duan,tnnls12:Duan}, the domain shift problem in ZSL is mainly due to the projection shift rather than the feature distribution shift.

   Many successful attempts have been made to address the domain shift problem in ZSL. A simple way is enlarging the seen data by employing a large amount of additional data with more categories and instances \cite{icip15:Xu}. The more abundant categories and data ensure the visual-semantic embedding better generalize to the unseen data. Another effective way is importance weighting, which is borrowed from the field of transfer learning \cite{is17:Ji}. For example, Xu \emph{et al.} \cite{eccv16:Xu} selectively re-weighted the relevant seen data to minimize the discrepancy between the marginal distributions of the seen and unseen data. The idea behind it is to augment the impacts of those data relevant to the unseen data to expect a better embedding generalization ability.

  Recently, there are considerable interests on developing elaborately transductive ZSL to rectify the domain shift problem. For example, Fu \emph{et al.} \cite{pami15:Fu} proposed a novel transductive multi-view hypergraph label propagation (TMV-HLP) model, in which the manifold structure of the unseen data is exploited to compensate for the impoverished supervision available from the sparse semantic vector. In this way, ZSL is achieved by semi-supervised label propagation from the semantic vector to the unseen data points within and across the graphs. With the idea that the instances of each class are condensed in a cluster in the deep feature space, \cite{arXiv16:Shojaee} first trained a linear transformation to map the class semantic vectors to the deep visual space, and then used a clustering algorithm to assign labels to instances of unseen classes. Under the transductive setting, \cite{iccv15:Kodirov} used the label embeddings of the unseen data to regularize the learned unseen domain projection under a sparse coding framework. Different from them, TASTE firstly uses the learned visual-semantic embedding from the seen data to predict the labels of unseen data, and then refines the learned visual-semantic embedding in an iterative way, where the classification capacity is progressively reinforced with highly reliable instances.

  \section{Proposed Approach}
    In this section, we first propose an Adaptive STructural Embedding (ASTE) model with the labeled seen data to learn the visual-semantic embedding, and then a Self-PAced selective Strategy (SPASS) is presented to alleviate the domain shift problem by gradually exploiting the potential label information in the unseen data in an easy to hard fashion. Finally, a Transductive ASTE (TASTE) method is developed to reinforce the discriminant capacity by combining the SPASS and ASTE.


\subsection{ASTE for ZSL}

    Let $\mathcal{S}=\{(\mathbf{x}_n,\mathbf{y}_n),n = 1,...,N\}$ denote the seen data from $K$ seen categories, where $\mathbf{x}_n\in\mathcal{X}$ is the input visual feature, $\mathbf{y}_n\in\mathcal{Y}$ is its label, $N$ is the number of the seen data. If $\mathbf{x}_n$ belongs to a class $k$, its corresponding class label is $\mathbf{y}_{n}=\mathbf{1}_{k}$ ($\mathbf{1}_{k}$ denotes a column vector of length $K$ with all zeros except a single $1$ at its $k$-th entry). We aim at learning $f: \mathcal{X}\rightarrow\mathcal{Y}$ between the input visual space $\mathcal{X}$ and the output structural label space $\mathcal{Y}$ by minimizing the empirical risk on the seen data $\frac{1}{N}\sum_{n=1}^{N}\ell(\mathbf{y}_n,f(\mathbf{x}_n))$, where $\ell(\mathbf{y}_n,f(\mathbf{x}_n))$ is the cost of predicting $f(\mathbf{x}_n)$ when its true label vector is $\mathbf{y}_n$. TABLE \uppercase\expandafter{\romannumeral1} shows the main notations used in this paper.

    We define a compatibility function $F : \mathcal{X}\times\mathcal{Y}\rightarrow\mathbb{R}$ to measure how compatible the pair $(\mathbf{x}_n,\mathbf{y}_n)$ are and formulate it as:
    \begin{equation}\label{1}
     F(\mathbf{x}_n,\mathbf{y}_n; \mathbf{W}) = g(\mathbf{x}_n;\mathbf{W})\bullet{\mathbf{y}_n},
    \end{equation}
     where the operation sign $\bullet$ denotes inner product, $g(\mathbf{x}_n;\mathbf{W}) = \mathbf{W}^{T}\mathbf{x}_n$ is the predicted label embedding of $\mathbf{x}_n$ in the seen label space, where $\mathbf{W} = [\mathbf{w}_1,...,\mathbf{w}_K]$ is the parameter matrix. The parameter vector of category $k$ is $\mathbf{w}_k$, which maps the input visual feature $\mathbf{x}_n$ to its compatibility score over class $k$.
     In this way, the function $F$ can be written as a bilinear form, i.e.,
    \begin{equation}\label{1}
        F(\mathbf{x}_n,\mathbf{y}_n;\mathbf{W}) = \mathbf{x}_n^T\mathbf{W}\mathbf{y}_n.
    \end{equation}

    Here the value of $F$ represents the compatibility score between the input visual feature and the output class label. The larger the value is, the more confidently that $\mathbf{x}_n$ belongs to class $\mathbf{y}_n$. Thus, the label prediction for an instance $\mathbf{x}_n$ is achieved by maximizing $F$ over the seen classes:
     \begin{equation}\label{1}
        f(\mathbf{x}_n;\mathbf{W}) = \arg\max_{\mathbf{y}_n\in\mathcal{Y}}F(\mathbf{x}_n,\mathbf{y}_n;\mathbf{W}),
    \end{equation}
    where $f(\mathbf{x}_n;\mathbf{W})$ is the predicted label of $\mathbf{x}_n$. It is the class label with the largest compatibility score.

\begin{table}
    \label{Table.1}
    \caption{\upshape The main notations.}
    \centering
    \begin{tabular}{c|c}
    \hline
    Notation & Description\\
    \hline
    $\mathcal{S}$  & seen data\\
    $\mathcal{U}$  & unseen data\\
    $\mathcal{X}$  & visual space\\
    $\mathcal{Y}$  & label space for seen classes\\
    $\mathcal{Z}$  & label space for unseen classes\\
    $K$  & number of the seen classes\\
    $N$  & number of the seen instances\\
    $L$  & number of the unseen classes\\
    $M$  & number of the unseen instances\\
    $C$  & trade-off parameter\\
    $p$  & dimensionality of visual space\\
    $q$  & dimensionality of class semantic embedding space\\
    $\xi_n$ & penalty term\\
    $\gamma$ & separation margin\\
    $\eta_t$ & learning step size\\
    $\theta$ & threshold parameter\\
    $\mathbf{V}\in{\mathbb{R}^{{p}\times{q}}}$ & compatibility matrix\\
    $\mathbf{A}_{s}\in{\mathbb{R}^{{q}\times{K}}}$ & class semantic matrix of seen classes\\
    $\mathbf{A}_{t}\in{\mathbb{R}^{{q}\times{N}}}$ & class semantic matrix of unseen classes\\
    $\mathbf{W}_s\in{\mathbb{R}^{p\times{K}}}$ &parameter matrix for seen classes\\
    $\mathbf{W}_t\in{\mathbb{R}^{p\times{L}}}$ &parameter matrix for unseen classes\\
    $F(\mathbf{x},\mathbf{y};\mathbf{W})$ & compatibility score of $\mathbf{x}$ over class $\mathbf{y}$\\
    $f(\mathbf{x},\mathbf{A};\mathbf{V})$ & predicted label of $\mathbf{x}$\\
    $g(\mathbf{x},\mathbf{A};\mathbf{V})$ & predicted vector of $\mathbf{x}$ in the label space\\
    \hline
    \end{tabular}
    \end{table}

    To achieve the knowledge transfer, we assume that the parameter vector of each category can be derived from its class semantic vector since it provides the corresponding class' properties. Based on this assumption, a function that independent from the categories is employed for obtaining the parameter vector $\mathbf{w}$ from the corresponding class semantic vector $\mathbf{a}$, i.e., $\mathbf{w}=\mathbf{V}\mathbf{a}$, where $\mathbf{V}\in\mathbb{R}^{p\times{q}}$ is the compatibility matrix shared by both the seen and unseen categories. To this end, the compatibility function for seen categories is written:
     \begin{equation}\label{1}
        F(\mathbf{x}_n,\mathbf{y}_n;\mathbf{W}_s) = \mathbf{x}_n^T\mathbf{W}_s\mathbf{y}_n=\mathbf{x}_n^T\mathbf{V}\mathbf{A}_s\mathbf{y}_n,
    \end{equation}
    where $\mathbf{W}_s = \mathbf{V}\mathbf{A}_s$ is the parameter matrix for seen categories and $\mathbf{A}_s=[\mathbf{a}_1,...,\mathbf{a}_K]\in\mathbb{R}^{q\times{K}}$ is the class semantic matrix for seen categories. Thus, $f(\mathbf{x}_n;\mathbf{W}_s)$ and $g(\mathbf{x}_n;\mathbf{W}_s)$ can also be written as $f(\mathbf{x}_n,\mathbf{A}_s;\mathbf{V})$ and $g(\mathbf{x}_n,\mathbf{A}_s;\mathbf{V})$, respectively.



    Now, the key problem is to learn the compatibility matrix $\mathbf{V}$ with the labeled seen data. In ASTE, the compatibility matrix is learned by enforcing the compatibility score of each seen instance over its correct class label to be ranked higher than that of the other classes, which can be formulated as follows:
     \begin{equation}\label{1}
         \max_{\mathbf{y}\in\mathcal{Y}\backslash{\mathbf{y}_n}}F(\mathbf{x}_n,\mathbf{y};\mathbf{W}_s)<{F(\mathbf{x}_n,\mathbf{y}_n;\mathbf{W}_s)}, ~~ \forall{n}\in[1,...,N].
    \end{equation}

    This formulation is closely related to \cite{cvpr13:Akata, cvpr15:Akata} and \cite{nips13:Frome}. Particularly, by formulating the visual feature and class semantic vector in a bilinear function, ALE \cite{cvpr13:Akata} and DeVISE \cite{nips13:Frome} use a pairwise ranking objective and SJE \cite{cvpr15:Akata} employs the unregularized structured SVM formulation. In contrast, ASTE enforces the compatibility score of each seen instance over its correct class to be ranked higher than the closest runner-up in a multiclass objective.



  According to the maximum-margin principle employed in SVM, we select the compatibility matrix $\mathbf{V}$ that makes the separation margin $\gamma$ (the minimal difference between the score of the correct label and the closest runner-up) be maximal.
  Restricting the $\ell_2$ norm of $\mathbf{w}$ to make the problem well-posed leads to the following optimization problem:
    \begin{equation}\label{1}
        \begin{aligned}
        &\max_{\gamma,\mathbf{w}_{l(\mathbf{x}_n)}: \|\mathbf{w}_{l(\mathbf{x}_n)}\|=1} \gamma \\
        &\emph{s.t.}~\mathbf{y}\in\mathcal{Y}\backslash{\mathbf{y}_n}:F(\mathbf{x}_n,\mathbf{y}_n;\mathbf{W}_s)-F(\mathbf{x}_n,\mathbf{y};\mathbf{W}_s)>\gamma,\\
        &~~~~\forall{n}\in[1,...,N],
        \end{aligned}
    \end{equation}
    where $\mathbf{w}_l(\mathbf{x}_n)$ is the parameter vector of the class that $\mathbf{x}_n$ belongs to, $\gamma = F(\mathbf{x}_n,\mathbf{y}_n;\mathbf{W}_s)-\max_{\mathbf{y}\in\mathcal{Y}\backslash{\mathbf{y}_n}}F(\mathbf{x}_n,\mathbf{y};\mathbf{W}_s)$.
   The problem can be equivalently expressed as a convex quadratic problem in the standard form:
   \begin{equation}\label{1}
        \begin{aligned}
        &\min \frac{1}{2}\|\mathbf{w}_{l(\mathbf{x}_n)}\|^2 \\
        &\emph{s.t.}~F(\mathbf{x}_n,\mathbf{y}_n;\mathbf{W}_s)-\max_{\mathbf{y}\in\mathcal{Y}\backslash{\mathbf{y}_n}}F(\mathbf{x}_n,\mathbf{y};\mathbf{W}_s)>1,\\
        & ~~~\forall{n}\in[1,...,N].
        \end{aligned}
    \end{equation}
 \begin{figure}[t]
    \begin{center}
   \includegraphics[height=4.5cm,width=7cm]{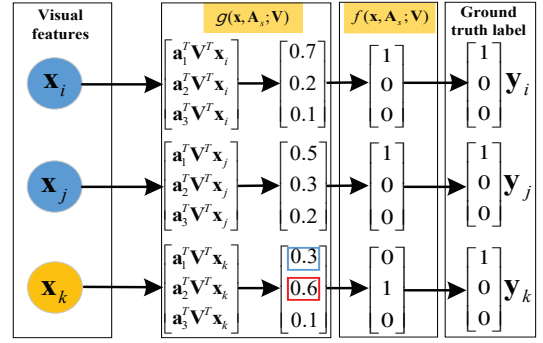}
    \end{center}
     \caption{Illustration of the principle of the cost function in the proposed ASTE. Suppose that there are three seen classes whose class semantic vectors are $\mathbf{a}_1$, $\mathbf{a}_2$ and $\mathbf{a}_3$; and $\mathbf{x}_i$, $\mathbf{x}_j$ and $\mathbf{x}_k$ are from the same class in this example. $\mathbf{x}_i$ and $\mathbf{x}_j$ are predicted correctly, while $\mathbf{x}_k$ is predicted incorrectly. For $\mathbf{x}_i$ and $\mathbf{x}_j$, the costs are only from the difference between $g(\mathbf{x},\mathbf{A}_s; \mathbf{V})$ and the ground truth label, they are 0.14 and 0.38, respectively. The cost of $\mathbf{x}_i$ is smaller than that of $\mathbf{x}_j$ since $\mathbf{x}_i$ is more reliable to be correctly-predicted than $\mathbf{y}_j$ (0.7 vs. 0.5). For $\mathbf{x}_k$, its cost is 1.16, which is from two parts: one is the difference between the largest compatibility score (the value in the red box, 0.6) and its ground truth compatibility score (the value in the blue box, 0.3), the other is the difference between $g(\mathbf{x},\mathbf{A}_s; \mathbf{V})$ and the ground truth label.}
    \label{fig:long}
    \label{fig:onecol}
  \end{figure}

    To allow the errors in the seen data, we add a penalty term to the objective function and relax the constraints:
    \begin{equation}\label{1}
        \begin{aligned}
        &\min \frac{C}{2}\|\mathbf{V}\mathbf{A}_s\|_F^2+\sum_n^{N}\xi_n \\
        &\emph{s.t.}~ \xi_n\geq1+F(\mathbf{x}_n,f(\mathbf{x}_n,\mathbf{A}_s;\mathbf{V}))-F(\mathbf{x}_n,\mathbf{y}_n;\mathbf{W}_s),\\
        &\forall{n}\in[1,...,N],~~ \xi_n\geq0,
        \end{aligned}
    \end{equation}
    where $C>0$ is a constant that controls the trade-off between the training error minimization and the margin maximization terms, $\xi_n$ is the slack variable, which can be viewed as the punishment for the instance $\mathbf{x}_n$ when it violates the constraint in Eq.~(5). For $\mathbf{x}_n$, if its predicted class label is the ground-truth label, i.e., $f(\mathbf{x}_n,\mathbf{A}_s;\mathbf{V}) = \mathbf{y}_n$, its cost is a constant $1$, otherwise, its cost is $1+F(\mathbf{x}_n,f(\mathbf{x}_n,\mathbf{A}_s;\mathbf{V}))-F(\mathbf{x}_n,\mathbf{y}_n;\mathbf{W}_s)$.

     We can observe that two different correctly-predicted instances are under equal punishments. Meanwhile, the incorrectly-predicted instances are under same punishment level against those correctly-predicted instances if the value of $F(\mathbf{x}_n,f(\mathbf{x}_n,\mathbf{A}_s;\mathbf{V}))-F(\mathbf{x}_n,\mathbf{y}_n;\mathbf{W}_s)$ is small. Clearly, this is unfair and cannot reflect the instances' differences. Intuitively, the correctly-predicted instances with higher compatibility scores should be imposed smaller punishments than those with smaller compatibility scores. Similarily, the incorrectly-predicted instances should be imposed more severe punishments than the correctly-predicted instances.

     Based on this assumption, we propose an adaptive function $\Delta: \mathcal{Y}\times\mathcal{Y}\rightarrow\mathbb{R}$ for differing the reliableness of the prediction. Specifically, the Euclidean distance between the predicted vector of an instance in the label space and its ground truth label is applied to define the adaptive function:
     \begin{equation}
        \Delta(\mathbf{y}_n,g(\mathbf{x}_n, \mathbf{A}_s;\mathbf{V})) = \|g(\mathbf{x}_n, \mathbf{A}_s;\mathbf{V})-\mathbf{y}_n\|_2^2,
    \end{equation}
     where $g(\mathbf{x}_n, \mathbf{A}_s;\mathbf{V})=(\mathbf{V}\mathbf{A}_s)^{T}\mathbf{x}_n$ is the predicted vector of $\mathbf{x}_n$ in the label space, $\Delta(\mathbf{y}_n,g(\mathbf{x}_n, \mathbf{A}_s;\mathbf{V}))$ quantifies the loss associated with $g(\mathbf{x}_n, \mathbf{A}_s;\mathbf{V})$ and its ground truth label. For descriptive convenience, we represent $\Delta(\mathbf{y}_n,g(\mathbf{x}_n, \mathbf{A}_s;\mathbf{V}))$ as $\Delta$ for short. The smaller the value of $\Delta$ is, the more confidently that the instance is predicted. An example is provided to explain the principle of cost function in ASTE in Fig. 2.

     Replacing the fixed margin in Eq.~(8) with $\Delta$, the final objective function of ASTE is obtained:
    \begin{equation}\label{1}
        \begin{aligned}
        &\min \frac{C}{2}\|\mathbf{V}\mathbf{A}_s\|_F^2+\sum_n^{N}\xi_n \\
        &\emph{s.t.}~ \xi_n\geq{\Delta}+F(\mathbf{x}_n,f(\mathbf{x}_n,\mathbf{A}_s;\mathbf{V}))-F(\mathbf{x}_n,\mathbf{y}_n;\mathbf{W}_s),\\
        &~~~\forall{n}\in[1,...,N],~~ \xi_n\geq0.
        \end{aligned}
    \end{equation}

    This problem can be equivalently expressed as a concave-convex program in the standard form:
      \begin{equation}\label{1}
        \begin{aligned}
        &\min \frac{C}{2}\|\mathbf{V}\mathbf{A}_s\|_F^2+\\
        &\sum_n^{N}\Delta+F(\mathbf{x}_n,f(\mathbf{x}_n,\mathbf{A}_s;\mathbf{V}))-F(\mathbf{x}_n,\mathbf{y}_n), \\
        &\forall{n}\in[1,...,N].
        \end{aligned}
    \end{equation}

    This objective function is similar to the latent structural SVM formulation \cite{cvpr10: Zhu, pami14:Xu}, which can be viewed as minimizing the sum of a convex and a concave function. It has been shown to converge to a local minimum or a saddle point solution \cite{nips09:Sriperumbudur}. The optimal compatibility matrix $\mathbf{V}^{*}$ can be obtained by a concave-convex procedure (CCCP), which consists in pair $(\mathbf{x}_n,\mathbf{y}_n)$ at each iteration and searches for class label $f(\mathbf{x}_n,\mathbf{A}_s;\mathbf{V})$ that achieves the highest compatibility score for $\mathbf{x}_n$. Specifically, if $f(\mathbf{x}_n,\mathbf{A}_s;\mathbf{V}) = \mathbf{y}_n$, the cost function for $\mathbf{x}_n$ is:
    \begin{equation}\label{1}
        \begin{aligned}
        \ell(\mathbf{x}_n) = \frac{C}{2N}\|\mathbf{V}\mathbf{A}_s\|_F^2+\Delta.
        \end{aligned}
    \end{equation}

    It is equivalent to the variation of ESZSL \cite{icml15:Romera-Paredes}, of which $\Delta$ is the loss function, and $\|\mathbf{V}\mathbf{A}_s\|_F^2$ is one regularizer. In this situation, ESZSL \cite{icml15:Romera-Paredes} can be seen as a special case of ASTE.
     If $f(\mathbf{x}_n,\mathbf{A}_s;\mathbf{V}) \neq \mathbf{y}_n$, the cost function for $\mathbf{x}_n$ is:
      \begin{equation}\label{1}
        \begin{aligned}
        &\ell(\mathbf{x}_n) = \frac{C}{2N}\|\mathbf{V}\mathbf{A}_s\|_F^2+\Delta+\\
        &~~~~~F(\mathbf{x}_n,f(\mathbf{x}_n,\mathbf{A}_s;\mathbf{V}))-F(\mathbf{x}_n,\mathbf{y}_n;\mathbf{W}_s).
        \end{aligned}
    \end{equation}

    The compatibility matrix $\mathbf{V}$ is updated by the stochastic gradient descent approach as follows:
       \begin{equation}\label{1}
        \begin{aligned}
        \mathbf{V}^{(t)} = \mathbf{V}^{(t-1)}-\eta_t{\frac{\partial\ell(\mathbf{x}_n)}{\partial{\mathbf{V}^{(t-1)}}}},
        \end{aligned}
    \end{equation}
    where $\eta_t$ is the learning step size at iteration $t$. The whole optimization procedure for Eq.~(11) is outlined in Algorithm 1.
    \begin{algorithm}
        \caption{The CCCP algorithm for solving Eq.~(11)}
        \KwIn{Seen data $\mathcal{S} =\{(\mathbf{x}_n,\mathbf{y}_n), 1\leq{n}\leq{N}\}$,\\
         ~~~~~~~~~the seen class semantic matrix $\mathbf{A}_s$,\\
         ~~~~~~~~~the initialized compatibility matrix $\mathbf{V}^{(0)}$,\\  
         ~~~~~~~~~the tolerance value $\epsilon$,\\
         ~~~~~~~~~the trade-off parameter $C$,\\
         ~~~~~~~~~the learning step size $\eta_t$.}
        \KwOut{The optimal compatibility matrix $\mathbf{V}^{*}$.}
    ~~~~1: $t$~~$\leftarrow$~{0},\\
    \textbf{Repeat}

    ~~~~2: Replace the $\mathbf{V}$ with $\mathbf{V}^{(t)}$, \\
    ~~~~3: Predict the class label $f(\mathbf{x}_n,\mathbf{A}_s;\mathbf{V})$ of a specific\\
    ~~~~~~ instance $\mathbf{x}_n$ by Eq.~(3),\\
    ~~~~4: Calculate the adaptive function $\Delta$ by Eq.~(9),\\
    ~~~~5: \textbf{if} $f(\mathbf{x}_n,\mathbf{A}_s;\mathbf{V}) = \mathbf{y}_n$\\
        ~~~~~~~~~ $\ell(\mathbf{x}_n) = \frac{C}{2N}\|\mathbf{V}\mathbf{A}_s\|_F^2+\Delta,$ \\
        ~~~~~~~\textbf{else}\\
        ~~~~~~~~~ $\ell(\mathbf{x}_n) = \frac{C}{2N}\|\mathbf{V}\mathbf{A}_s\|_F^2+\Delta+F(\mathbf{x}_n,f(\mathbf{x}_n,\mathbf{A}_s;\mathbf{V}))-F(\mathbf{x}_n,\mathbf{y}_n),$\\
    ~~~~6: $t$~~$\leftarrow$~~{$t+1$},\\
    ~~~~7: Update $\mathbf{V}^{(t)}$ by the stochastic gradient descent\\
    ~~~~~~ approach by Eq.~(14).\\

    \textbf{until} {\textit{The loss in Eq.~(11) is under the tolerance $\epsilon$.}}
     \end{algorithm}

Once the optimal compatibility matrix $\mathbf{V}^{*}$ is obtained, the class label of an unseen instance $\mathbf{x}_m$ can be predicted by resorting to the largest compatibility score that corresponds to one of the unseen categories:
\begin{equation}
    \mathbf{z}^{\ast} =  \arg\max_{\mathbf{z}\in\mathcal{Z}}\mathbf{x}_m^{T}\mathbf{V}\mathbf{A}_t\mathbf{z},
\end{equation}
where $\mathcal{Z}$ is the label space of unseen classes that is disjoint from that of the seen class $\mathcal{Y}$, i.e., $\mathcal{Y}\cap\mathcal{Z} = \emptyset$, and $\mathbf{A}_t$ is the unseen class semantic matrix in which each column denotes a class' semantic vector. The flowchart of the proposed ASTE is illustrated in Fig.~3.

  \begin{figure}[t]
    \begin{center}
   \includegraphics[height=4.0cm,width=8.8cm]{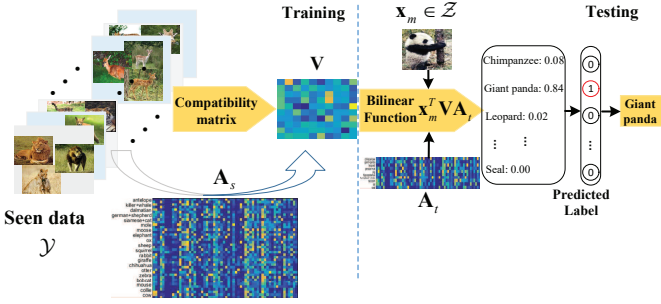}
    \end{center}
     \caption{The flowchart of the proposed ASTE for ZSL. At the training stage, the seen data and its class semantic matrix $\mathbf{A}_s$ are used for training the compatibility matrix $\mathbf{V}$, which is applied for predicting the unseen data. At the testing stage, a testing instance from the unseen label space $\mathcal{Z}$ that is disjoint from the seen label space $\mathcal{Y}$ is predicted by resorting to the highest compatibility score of the testing instance over the unseen categories that corresponds to.}
    \label{fig:long}
    \label{fig:onecol}
  \end{figure}

\subsection{The Self-PAced Selective Strategy (SPASS)}

Most inductive ZSL approaches directly apply the learned model from the seen data to recognize the unseen data. However, the different distribution between the seen data and the unseen data will lead the learned embedding model to be biased on the unseen data. In order to address this domian shift issue, we take the ZSL as a special case of transductive learning. Considering that the primary cause of domain shift problem in ZSL lies in the label absence of the unseen data, a natural idea is to select some pseudo labeled instances from unseen data in a transductive learning way to tackle it. Thus, the challenge is how to choose the reliable pseudo labeled data. To address this issue, we propose a selective strategy to gradually select the unseen data in an easy to hard fashion. As defined in \cite{icml09:Bengio}, the process of gradually added training instances is called a curriculum. A straightforward way to design a curriculum is to select the instances based on certain heuristical ``easiness" measurements. In the curriculum, the easy instances are selected for training. Based on this idea, the previously predicted unseen data are ranked according to the compatibility scores. The higher compatibility score is, the more reliable of an instance to be correctly predicted. In this paper, the easiness of an instance is defined as: an instance is easy if it is reliable to be predicted correctly.



To this end, we present a Self-PAced Selective Strategy (SPASS) to iteratively select the reliable pseudo labeled unseen instances. Inspired by self-paced learning \cite{nips10:Kumar}, a binary variable $u_m$ is introduced to indicate whether the $m$-th instance is easy or not. Then, the selective process is defined by:
\begin{equation}\label{4}
    \begin{aligned}
         \arg\min_{\mathbf{u}}\sum_{m=1}^{M}u_m\|f(\mathbf{x}_m;\mathbf{W}_t) - g(\mathbf{x}_m;\mathbf{W}_t)\|_F^2-\theta{u_m},
    \end{aligned}
\end{equation}
where $M$ is the instance number of the unseen data, $f(\mathbf{x}_m;\mathbf{W}_t)$ is the predicted class label of $\mathbf{x}_m$ by Eq.~(3),  and $g(\mathbf{x}_m;\mathbf{W}_t)=\mathbf{W}_t^{T}\mathbf{x}_m$ is the embedding of the test instance $\mathbf{x}_m$ in the unseen label space, $\mathbf{W}_t=\mathbf{V}\mathbf{A}_t$ is the parameter matrix for unseen classes; $\mathbf{u}=[u_1,...,u_M]\in[0,1]^{M}$ is the indicate vector, and $\theta$ is a threshold parameter for controlling the number of instances to be selected. $\|f(\mathbf{x}_m;\mathbf{W}_t) - g(\mathbf{x}_m;\mathbf{W}_t)\|_F^2$ is the potential loss for unseen instance $\mathbf{x}_m$. The smaller it is, the more reliable an instance is correctly predicted. In this way, the instances whose losses are smaller than a certain threshold $\theta$ are taken as ``easy" instances, and will be selected as the pseudo labeled data, otherwise unselected. When $\theta$ is small, only ``easy" instances are selected. As $\theta$ grows, more instances with larger losses will be gradually aggregated with the labeled set for a more suitable model. The selective process stops when all unseen data are selected.

\subsection{Transductive ASTE (TESTE)}
 In the situation of both the labeled seen data and the unlabeled unseen data are off-the-shelf, a transductive ZSL approach can be developed by combining the SPASS and ASTE. We call it Transductive ASTE (TASTE). The united framework of TASTE is:
\begin{equation}\label{4}
    \begin{aligned}
 &\arg\min_{\mathbf{V}}\sum_n^{N}\mathcal{L}(\mathbf{y}_n,f(\mathbf{x}_n,\mathbf{A}_s;\mathbf{V}))\\
 &+\sum_m^{M}\mathcal{L}(f(\mathbf{x}_m,\mathbf{A}_t;\mathbf{V}),g(\mathbf{x}_m,\mathbf{A}_t;\mathbf{V}))
 +\lambda{\mathcal{R}},
    \end{aligned}
\end{equation}
where $\mathcal{L}(\centerdot)$ is the loss function, $\mathcal{R}$ is a regularization function, and $\lambda$ is the trade-off weight. Specifically,

\begin{equation}\label{4}
    \begin{aligned}
        \mathcal{L}(\mathbf{y}_n,f(\mathbf{x}_n,\mathbf{A}_s;\mathbf{V}))=\Delta(\mathbf{x}_n,f(\mathbf{x}_n,\mathbf{A}_s;\mathbf{V}))\\
             ~~+F(\mathbf{x}_n,f(\mathbf{x}_n,\mathbf{A}_s;\mathbf{V}))-F(\mathbf{x}_n,\mathbf{y}_n),
    \end{aligned}
\end{equation}
\begin{equation}\label{4}
 \begin{aligned}
        &\mathcal{L}(f(\mathbf{x}_m,\mathbf{A}_t;\mathbf{V}),g(\mathbf{x}_m,\mathbf{A}_t;\mathbf{V}))=\\
        &~~~~~~~~~u_m\|f(\mathbf{x}_m,\mathbf{A}_t;\mathbf{V})-g(\mathbf{x}_m,\mathbf{A}_t;\mathbf{V})\|_2^2,
 \end{aligned}
\end{equation}

\begin{equation}\label{4}
 \begin{aligned}
        \mathcal{R} = \|\mathbf{V}\mathbf{A}_s\|_F^2+\|\mathbf{V}\mathbf{A}_t\|_F^2.
 \end{aligned}
\end{equation}




The solution to Eq.~(17) is addressed by the Alternative Convex Search (ACS) method \cite{inc93:Bazaraa} in which the variables to be optimized (i.e., $\mathbf{u}$ and $\mathbf{V}$) are divided into two disjoint blocks. When $\mathbf{u}$ is fixed, the problem converts a convex and concave function. Thus, $\mathbf{V}$ can be obtained by the CCCP algorithm. More specifically, for labeled pair $(\mathbf{x}_n,\mathbf{y}_n)$, its loss is obtained by Eq. (13). As for the unseen instance $\mathbf{x}_m$, the predicted label $f(\mathbf{x}_m,\mathbf{A}_t;\mathbf{V})$ is taken as its pseudo label, of which the easiness is controlled by the value of $u_m$. The loss for it is:
  \begin{equation}\label{1}
        \begin{aligned}
        &\ell(\mathbf{x}_m) = \frac{C}{2M}\|\mathbf{V}\mathbf{A}_t\|_F^2+\\
        &~~~~~~~~u_m\|f(\mathbf{x}_m,\mathbf{A}_t;\mathbf{V}) - g(\mathbf{x}_m,\mathbf{A}_t;\mathbf{V})\|_2^2.
        \end{aligned}
  \end{equation}

Minimizing this loss function means to enforce the embedding of unseen instance to be close to its predicted class label. In this way, the $\mathbf{V}$ is updated by Eq.~(14).

On the other hand, fixing $\mathbf{V}$, then $\mathbf{u}$ can be easily updated by:
\begin{equation}
u_m =
\left\{
             \begin{array}{lr}
             1, ~\|f(\mathbf{x}_m,\mathbf{A}_t;\mathbf{V}) - g(\mathbf{x}_m,\mathbf{A}_t;\mathbf{V})\|_2^2~<~\theta, \\
             0, ~otherwise.
             \end{array}
\right.
\end{equation}

It means that a specific instance $\mathbf{x}_m$ whose loss is smaller than a certain threshold $\theta$ is selected as ``easy" instance, and its pseudo label is $f(\mathbf{x}_m,\mathbf{A}_t;\mathbf{V})$; otherwise, it is unselected.
As $\theta$ grows, more reliable instances will be gradually added into the labeled set, thus the classification capacity is progressively reinforced.

\section{A novel Fast Training (FT) strategy}
  The approaches with closed-form solution, such as ESZSL \cite{icml15:Romera-Paredes} and Linear Regression (LR) \cite{ecml15:Shigeto}, generally have a relatively fast training speed. In contrast, the gradient descent based approaches, such as SJE \cite{cvpr15:Akata}, DeViSE \cite{nips13:Frome},  ASTE and TASTE, are slow due to the heavy iterative computational burdens. In order to improve the training efficiency, we propose a surprisingly simple fast training (FT) strategy by applying the visual pattern of each seen class to represent the corresponding class. In this paper, the visual pattern of a class is defined as the average visual features in a class. The idea behind it is that one class can be well represented as a semantic vector, thus we can try to represent each class with its visual pattern in the visual space. In this way, the FT strategy simply takes the visual pattern of each class as training data which greatly alleviates the computational burdens especially for those gradient descent based approaches.


  Simple as FT strategy is, it is effective. The experiments show it can speed up the running time of most existing methods by 4$\thicksim$300 times while holding the previous performance. However, it is hard to explain this interesting phenomenon, i.e., the performance is almost steady with and without FT strategy. An intuitive explanation is that the FT strategy can suppress the noise and particularly possible outliers during the adaptation. Besides, we also try to provide its theoretical analysis in the risk bounds for domain adaptation in the following.

  Generally, the objective functions of different domain adaptation approaches on the seen data can be expressed as:
  \begin{equation}\label{5}
    \begin{aligned}
   \arg\min\sum_{\mathbf{x}_n\in\mathcal{S}}\mathcal{L}(\mathbf{y}_n;f(\mathbf{x}_n; \mathbf{W}_s))+\mathcal{R},
    \end{aligned}
  \end{equation}
  where $\mathbf{y}_n$ and $f(\mathbf{x}_n; \mathbf{W}_s)$ are ground truth class label and the predicted class label, respectively.
  The expected error in the seen data $\mathcal{S}$ and the unseen data $\mathcal{U}$ are defined as:
  \begin{equation}\label{6}
    \begin{aligned}
        &\epsilon_s(f) = \mathbb{E}_{\mathbf{x}_n\thicksim\mathcal{S}}[|f(\mathbf{x}_n;\mathbf{W}_s)-\mathbf{y}_n|],\\
        &\epsilon_t(f) = \mathbb{E}_{\mathbf{x}_m\thicksim\mathcal{U}}[|f(\mathbf{x}_m;\mathbf{W}_t)-\mathbf{y}_m|].\\
    \end{aligned}
  \end{equation}

According to the Theorem 1 in \cite{nips06:Ben-David}, given a hypothesis space $\mathcal{H}$ of VC-dimension $\bar{d}$ in $\mathcal{U}$ that contains $\bar{m}$ instances, then with probability at least $1-\delta$, for every $f\in\mathcal{H}$, its expected error in $\mathcal{U}$ is:
\begin{equation}\label{7}
    \begin{aligned}
        &\epsilon_t(f) \leq\epsilon_s(f)+ 4\sqrt{\frac{2\bar{d}}{\bar{m}}(\log\frac{2\bar{m}}{\bar{d}}+\log\frac{4}{\delta})}\\
       & ~~~~~~~~~~+\lambda+d_\mathcal{H}(\mathcal{S},\mathcal{U}),
    \end{aligned}
\end{equation}
where $\lambda$ is an upper-bound of $\inf_{f\in\mathcal{H}}[\epsilon_s(f)+\epsilon_t(f)]$, $d_\mathcal{H}(\mathcal{S},\mathcal{U})$ is the distribution distance between $\mathcal{S}$ and $\mathcal{U}$.

Here, the FT strategy mainly brings the variation on the first term $\epsilon_s(f)$ in Eq.~(25), which is the expected error in the seen data. It is obtained by summing the expected error of each instance. With FT strategy, only the expected error of the visual pattern in each seen class is considered. The value of $\epsilon_s(f)$ of the approaches with the FT strategy is smaller than those without FT strategy. This conclusion will be verified in the following. Then, the error bound $\epsilon_t(f)$ in Eq.~(25) is smaller than those approaches without FT strategy, which indicates the information is easier to transfer from the seen domain to the unseen domain. Therefore, the performances will be hold when applying the FT strategy to most existing methods.

Without loss of generality, we take the Linear Regression (LR) as an example to verify the above conclusion since LR is simple and has a closed-form solution.
Given the model parameter $\mathbf{W}$, the empirical loss of LR for $N$ seen instances from $K$ classes is:
\begin{equation}\label{8}
    \begin{aligned}
        \epsilon_s = \frac{1}{N}\sum_{i=1}^{K}\sum_{j=1}^{k_i}\|\mathbf{x}_{ij}^T\mathbf{W}-y_i\|_2^2,
    \end{aligned}
\end{equation}
where $\mathbf{x}_{ij}$ is the $j$-th instance from $i$-th class, $k_i$ is the instance number of class $i$.
In contrast, the empirical loss in LR with FT strategy is:
\begin{equation}\label{9}
    \begin{aligned}
      \bar{\epsilon}_s = \frac{1}{K}\sum_{i=1}^K\|\bar{\mathbf{x}}_i^T\mathbf{W}-y_i\|_2^2,
    \end{aligned}
\end{equation}
where $\bar{\mathbf{x}}_i= \frac{1}{k_i}\sum_{j=1}^{k_i}\mathbf{x}_{ij}$ is the visual pattern for class $i$. In Eq.~(26), the operation $\|\bullet\|$ is calculated with $N$ times, while it is calculated with $K$ times in Eq.~(27), in which $K\ll{N}$. Furthermore, we have proposition 1.\\
\textbf{Proposition 1.} For a specific class $i$, we can conclude
    \begin{equation}\label{10}
    \begin{aligned}
       \bar{\mathbf{x}}_i\bar{\mathbf{x}}_i^{T}\leq\frac{1}{k_{i}}\sum_{j=1}^{k_i}\mathbf{x}_{ij}\mathbf{x}_{ij}^T.
    \end{aligned}
\end{equation}

\emph{Proof:} For a specific matrix $\mathbf{X} = [\mathbf{x}_1,...,\mathbf{x}_m]$, its variance is:
\begin{equation}\label{11}
    \begin{aligned}
       D(\mathbf{X}) = E(\mathbf{x}_i^2)-[E(\mathbf{X})]^2,
    \end{aligned}
\end{equation}
where $E(\mathbf{X})=\frac{1}{m}\sum_i^{m}\mathbf{x}_i$ is the average of $\mathbf{X}$ and $E(\mathbf{x}_i^2) = \frac{1}{m}\sum_i^{m}\mathbf{x}_i\mathbf{x}_i^T$. It is easy to verify Eq.~(28) because the variance value $D(\mathbf{X})$ is non-negative.

  According to Proposition 1, we can conclude that the value of $\bar{\epsilon}_s$ is smaller than that of $\epsilon_s$ for LR. Replace the $\epsilon_s(f)$ with $\bar{\epsilon}_s(f)$ in Eq.~(25), the upper bound $\epsilon_t$ of the unseen domain decreases correspondingly, which verifies the conclusion that the expected error with FT strategy is smaller than that without FT strategy. 


\section{Experiments}
    To evaluate the effectiveness and efficiency of the proposed approaches, we conduct extensive experiments on three benchmark zero-shot learning datasets. We first detail the datasets and the experimental settings, then present the experimental results of ASTE, followed by the comparative results of TASTE. Finally, the evaluations of the FT strategy on ASTE, TASTE and other ZSL approaches are provided.
\subsection{Datasets and experimental settings}
    \textbf{Datasets.} The proposed approaches are evaluated on three benchmark datasets: Animal with Attributes (AwA) \cite{cvpr09:Lampert}, Caltech UCSD Birds (CUB) \cite{Technical11:Wah}, and aPascal-aYahoo (aPY) \cite{cvpr09:Farhadi}. AwA is a standard attribute dataset for ZSL; CUB is a fine-grained dataset with little variations among different classes; aPY is a combined dataset of aPascal and aYahoo, in which aYahoo dataset is collected from Yahoo image search that is different from the ones in aPascal. These datasets contain diverse categories such as animals, birds and objects. More specifically,
    AwA contains 30,475 images from 50 different animals, paired with a set of human provided 85 continuous attributes for each class. We follow the standard seen/unseen split \cite{cvpr09:Lampert}, where 40 classes with 24,295 images are taken as the seen data and the remaining 10 classes with 6180 images are adopted as the unseen data. CUB contains 11,788 images from 200 bird species with 312 associated attributes. In this dataset, we use the same zero-shot split as \cite{cvpr15:Akata} with 150 classes for seen data and 50 disjoint classes for the unseen data. For aPY, it contains 2,644 images from 32 classes, in which each image is annotated by 64 binary attributes. To represent each class with an attribute vector, we average the attributes of the images in each class. In the experiment, the aPascal is used as the seen data, and the aYahoo is used as the unseen data.

    \textbf{Experimental settings.} To make our approaches easily compared with the previous approaches, we use the popular VGG-VeryDeep-19 \cite{iclr14:Simonyan} model that pre-trained on imageNet to extract the visual features. In specific, we use the fully connected layer (FC7 layer of VGG-VeryDeep-19) for representing the image. We denote it with VGG in the following.

    There are two parameters in TASTE (the weight $C$ and the threshold $\theta$) and one parameter in ASTE (the weight $C$),  which can be determined via cross validation. However, in our experiments, we found the performance is not sensitive to the weight $C$, thus we fix $C = 0.1$. As for the threshold $\theta$, we initiate it with the half value of the maximal loss in the unseen data as the initial threshold, then gradually increase its value until it equals to the maximal loss in the unseen data. In this paper, $\theta$ is gradually selected from $(0.5\delta, 0.7\delta, 0.9\delta, \delta)$, where $\delta$ is the maximal value of the loss function.

    The optimization of $\mathbf{V}$ is performed with stochastic gradient descent approach, which is initialized randomly with normal distribution. The size of the mini-batch is 50, the updating rate $\eta_t$ is selected from $\{1,~0.1,~0.01\}$ successively with 50 times as a round. As the proposed ASTE is a non-convex problem, thus different initialization will lead to different local minimum. For a fair comparison, we perform 5 trials and report the mean and the standard variance as the final performance. For all datasets, we apply the popular average per-class top-1 accuracy to evaluate the performance \cite{cvpr09:Lampert,cvpr15:Fu,cvpr13:Akata,cvpr15:Akata,icml15:Romera-Paredes,cvpr16:Changpinyo}. The average testing time in our Matlab implementation is about 1ms per unseen instance in a desktop computer with an Intel Core i7-4790K processor and 32G RAM.

\subsection{Experimental results of ASTE}

\begin{table*}
\caption{\upshape Performance comparison (\%) of different inductive approaches on AwA, CUB and aPY datasets in form of mean $\pm$ standard variance. Notations: $\mathcal{G}$ and $\mathcal{V}$ represent features obtained from models of GoogleNet and VGG-verydeep-19, respectively. The mark \S ~indicates the comparative method is implemented by ourselves. We report their best performance after tuning the parameters in their models.}
\begin{center}
\begin{tabular}{|l|c|c|c|c|}
\hline
Feature &Method & AwA &CUB & aPY \\
\hline\hline
\multirow{3}{*}{~~~~$\mathcal{G}$} 
&LatEm \cite{cvpr16:Xian} & 71.9 & 45.5 & - \\
& Changpinyo \textit{et al.} \cite{cvpr16:Changpinyo} & 72.9 & \textbf{54.7} & - \\
& HAT \cite{wacv15:Al-Halah} & 74.9 & 51.8 & - \\
\hline
\multirow{8}{*}{~~~~$\mathcal{V}$} &DAP \cite{cvpr09:Lampert} & 57.23 & - & 38.16 \\
&KDICA \cite{arXiv16:Gan} & 73.8 & 43.7 & -\\
&SJE \cite{cvpr15:Akata}\S & 80.1$\pm$1.13 & 46.3$\pm$1.86 &44.5$\pm$1.62\\
&ESZSL \cite{icml15:Romera-Paredes}\S  &74.4$\pm$0.58 & 46.8$\pm$0.61 & 41.8$\pm$1.2 \\
&SSE-ReLU \cite{iccv15:Zhang} & 76.33$\pm$0.83 & 30.41$\pm$0.2 & 46.23$\pm$0.53 \\
&Zhang \textit{et al.} \cite{cvpr16:Zhang} & 80.46$\pm$0.53 & 42.11$\pm$0.55 & 38.94$\pm$2.27\\
&Bucher \textit{et al.} \cite{eccv16:Bucher} & 77.32$\pm$1.03& 43.29$\pm$0.38 & \textbf{53.15$\pm$0.88}\\
&ASTE  &\textbf{81.0$\pm$0.64} & 50.2$\pm$1.28 &48.0$\pm$1.96\\
\hline
\end{tabular}
\end{center}

\end{table*}

\textbf{Competitors}. Ten state-of-the-art inductive ZSL approaches are selected for comparison with ASTE, ranging from linear-based \cite{cvpr09:Lampert,iccv15:Zhang,cvpr16:Zhang}, nonlinear-based \cite{cvpr16:Changpinyo,arXiv16:Gan,wacv15:Al-Halah}, bilinear-based \cite{cvpr16:Xian,icml15:Romera-Paredes} and max margin-based \cite{cvpr15:Akata,eccv16:Bucher}. The performance results of the selected approaches are all from the original papers except \cite{cvpr15:Akata} and \cite{icml15:Romera-Paredes}. As for SJE \cite{cvpr15:Akata} and ESZSL \cite{icml15:Romera-Paredes}, we implement them with VGG features by ourselves. Specifically, the settings in SJE is similar to ASTE, and the parameters in ESZSL are picked via cross validation.

\textbf{Comparative Results}. The comparison results are outlined in Table \uppercase\expandafter{\romannumeral2}, which shows that ASTE achieves the state-of-the-art performance on three datasets. It is highlighted that the ASTE performs the best on AwA dataset. Specifically, with the same VGG features, ASTE outperforms DAP \cite{cvpr09:Lampert}, KDICA \cite{arXiv16:Gan}, SJE \cite{cvpr15:Akata}, ESZSL \cite{icml15:Romera-Paredes}, SSE-ReLU \cite{iccv15:Zhang}, Zhang \textit{et al.} \cite{cvpr16:Zhang}, Bucher \textit{et al.} \cite{eccv16:Bucher} in 23.77, 7.2, 0.9, 6.6, 4.67, 0.54 and 3.68 absolute percentage point on AwA dataset, respectively. On CUB dataset, ASTE achieves the third best performance, which is inferior to \cite{cvpr16:Changpinyo} and \cite{wacv15:Al-Halah} in 4.5\% and 1.6\%, respectively, and superior to KDICA \cite{arXiv16:Gan}, SJE \cite{cvpr15:Akata}, ESZSL \cite{icml15:Romera-Paredes}, SSE-ReLU \cite{iccv15:Zhang}, Zhang \textit{et al.} \cite{cvpr16:Zhang}, Bucher \textit{et al.} \cite{eccv16:Bucher}, in 6.5, 3.9, 3.4, 19.79, 8.09 and  6.91 absolute percentage point, respectively. On aPY dataset, ASTE achieves the second best performance, which has 9.84\%, 3.5\%, 6.2\%, 1.77\%, 9.06\% gains against DAP \cite{cvpr09:Lampert}, SJE \cite{cvpr15:Akata}, ESZSL \cite{icml15:Romera-Paredes}, SSE-ReLU \cite{iccv15:Zhang}, Zhang \textit{et al.} \cite{cvpr16:Zhang}, respectively.

\subsection{Experimental results of TASTE}

First, we compare in Fig.~4 the performance of ASTE and TASTE. It can be observed that the usage of SPASS improves the performance of ASTE significantly. Specifically, the performances have 8.74\% and 17.33\% improvements in AwA and aPY datasets, respectively. In contrast, the performance has a smaller improvement of 4.02\% in CUB dataset. This is mainly due to that the CUB dataset is a fine-grained dataset which is more difficult to be classified, thus SPASS may select some less reliable instances that provide little positive information for refining the model.

Next, the performance superiority of the TASTE is evaluated against five state-of-the-art transductive ZSL methods, as shown in Table \uppercase\expandafter{\romannumeral3}. It can be observed that TASTE outperforms the others on both AwA and aPY datasets, and is comparable to the others on CUB dataset. Specifically, the absolute performance gains of TASTE against the others are from 1.1\% to 16.54\% in AwA dataset. It is more significant on aPY dataset, which has noticeable improvements of 26.3\% and 15.56\% against SMS \cite{aaai16:Guo} and \cite{arXiv16:Shojaee}, respectively. On CUB dataset, TASTE is inferior to \cite{arXiv16:Shojaee} in 4.55\%, but outperforms the others significantly. Overall, the results on the three benchmark datasets demonstrate that the effectiveness of SPASS and TASTE.

\begin{figure}[t]
\begin{center}
   \includegraphics[height=4.3cm,width=6.8cm]{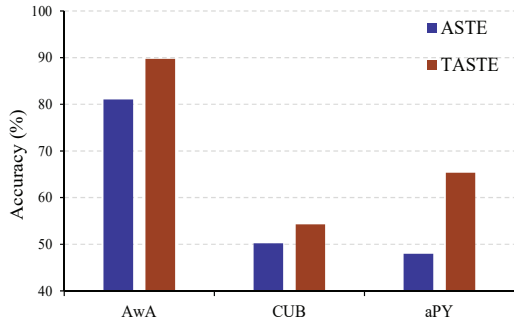}
\end{center}
   \caption{Comparative results of ASTE and TASTE on AwA, CUB and aPY datasets, respectively.}
\label{fig:long}
\label{fig:onecol}
\end{figure}

\begin{table}
\caption{\upshape Performance comparison (\%) of different transductive-based ZSL methods \protect \\ on different datasets.
 Notations: features $\mathcal{V}$, $\mathcal{O}$ and $\mathcal{D}$ represent VGG-verydeep-19, OverFeat and DeCAF, respectively.}
\begin{center}
\begin{tabular}{|l|c|c|c|c|}
\hline
Feature &Method & AwA & CUB & aPY \\
\hline\hline
~$\mathcal{O}$\&$\mathcal{D}$ & TMV-HLP \cite{pami15:Fu} &80.5 &47.9 &-\\
\hline
\multirow{2}{*}{~~~~$\mathcal{D}$}
&SMS \cite{aaai16:Guo}  &78.47 & - & 39.03\\
&Kodirov \textit{et al.} \cite{iccv15:Kodirov} &73.2 & 39.5 & -\\
\hline
\multirow{3}{*}{~~~~$\mathcal{V}$} &Shojaee \textit{et al.} \cite{arXiv16:Shojaee} & 88.64 & \textbf{58.80} & 49.77\\
&Wang \textit{et al.} \cite{arXiv16:Wang} & 87.9 & 53.5 & - \\
&TASTE & \textbf{89.74} & 54.25 & \textbf{65.33}\\
\hline
\end{tabular}
\end{center}
\end{table}
\subsection{Evaluation of the FT strategy}

To confirm the efficacy and efficiency of the proposed FT strategy, we conduct a set of experiments on the three benchmark datasets. Meanwhile, three additionally typical methods are chosen to indicate the generality of the FT strategy. These methods are LR \cite{ecml15:Shigeto}, ESZSL \cite{icml15:Romera-Paredes} and SJE \cite{cvpr15:Akata}, representing the linear approach, bilinear approaches with and without closed-form solutions, respectively. To facilitate the introduction, we add a prefix ¡®Fast-¡¯ to the name of these methods, and call them fast methods. Specifically, for Fast-SJE, Fast-ASTE, and Fast-TASTE, we initialize the compatibility matrix $\mathbf{V}$ with the optimal parameters in ESZSL to reach the convergence quickly, since ESZSL has a closed-form solution.

Table \uppercase\expandafter{\romannumeral4} depicts the performance comparisons of those with and without the FT strategy. It can be observed that the performances of the fast methods are close to that of their corresponding methods on all three datasets.
 In particular, all the fast methods outperform their corresponding ones in aPY dataset. In AwA dataset, the FT strategy improves those closed-form expression approaches LR and ESZSL and  slightly lower than those without closed-form expression approaches.
 For CUB dataset, except that Fast-ESZSL and Fast-TASTE are slightly lower than ESZSL and TASTE, the other methods also have an obvious performance improvement. In addition, from the perspective of the fast methods, all of them have performance improvements on at least two datasets except for TASTE. Therefore, the FT strategy is effective in holding the original methods' performance. On the other hand, almost all the standard variances of the fast methods are smaller than their corresponding original approaches, which indicates the fast methods are more robust to noises and particularly possible outliers.

Table \uppercase\expandafter{\romannumeral5} verifies the efficiency of the FT strategy. It can be seen that the fast methods speed up the run time significantly. Concretely, for those methods with closed-form solutions, i.e., Fast-LR and Fast-ESZSL, the speed up ratios are about 4. More impressively, the speed up ratios are about 230$\thicksim$310 for those methods without closed-form solutions, including Fast-SJE, Fast-ASTE, and Fast-TASTE.

\begin{table}
\caption{\upshape Performance comparison (\%) in the form of mean $\pm$ standard variance of different methods with and without the FT strategy. The mark $\uparrow$ indicates the performance improvement.}
\begin{center}
\begin{tabular}{|l|c|c|c|}
\hline
Method & AwA & CUB & aPY \\
\hline\hline
LR \cite{ecml15:Shigeto} & 65.0$\pm$1.67 & 40.2$\pm$1.34 & 40.2$\pm$1.92\\
Fast-LR & 67.5$\pm$0.27$\uparrow$ & 41.8$\pm$0.46 $\uparrow$ & 46.5$\pm$1.16$\uparrow$\\
\hline
ESZSL \cite{icml15:Romera-Paredes} &74.4$\pm$0.58 & 46.8$\pm$0.61 & 41.8$\pm$1.21\\
Fast-ESZSL  & 75.6$\pm$0.20$\uparrow$ & 46.5$\pm$0.43 & 48.6$\pm$1.16$\uparrow$ \\
\hline
SJE \cite{cvpr15:Akata} & 80.1$\pm$1.13 & 46.3$\pm$1.86 &44.5$\pm$1.62 \\
Fast-SJE & 77.1$\pm$0.57 & 48.8$\pm$0.55$\uparrow$ & 49.6$\pm$1.84$\uparrow$ \\
\hline
ASTE & 81.0$\pm$0.64 & 50.2$\pm$1.28 &48.0$\pm$1.96 \\
Fast-ASTE & 78.1$\pm$0.6 & 51.5$\pm$0.61$\uparrow$ &51.4$\pm$1.29$\uparrow$\\
\hline
TASTE & 89.7$\pm$0.51 &54.3$\pm$0.82 &65.3$\pm$1.86\\
Fast-TASTE & 88.9$\pm$0.26 & 54.1$\pm$0.55 & 66.5$\pm$1.01$\uparrow$\\
\hline
\end{tabular}
\end{center}

\end{table}


    What's more, we also evaluate the impact of the amount of training data to the visual-semantic embedding models. To this end, we first randomly select one instance from each class for training the model, and then vary the percent of each training class number from 10\% to 100\% in intervals of 10\% of the corresponding class. All results are reported with the average value of 5 trials. As Fig.~5 shows, with the increase of the training instances, the classification performances steadily increase for all approaches when the rate of each training class number is lower than 0.4. In contrast, when the percent of each training class number is above 0.4, the performances of different approaches reach a plateau, possibly because there is no further margin to improve \cite{icml15:Romera-Paredes}. Meanwhile, we can observe that the performance with randomly selective strategy (i.e., randomly choose one instance from each class) is worse than that of the FT strategy, which suggests that the success of FT strategy is due to the representative power of the all input data in each class.

\begin{table}
\caption{\upshape Training time of different approaches on AwA dataset with and without the FT strategy.}
\begin{center}
\begin{tabular}{|l|c|c|c|}
\hline
Method & withour FT & with FT & Ratio of speed up\\
\hline
LR \cite{ecml15:Shigeto} & 4.98s & 1.21s &4.11 \\
\hline
ESZSL \cite{icml15:Romera-Paredes} & 5.33s & 1.19s & 4.48 \\
\hline
SJE \cite{cvpr15:Akata} &11124.57s &35.03s&317.57 \\
\hline
ASTE & 12420.61s &50.12s&247.92\\
\hline
TASTE & 26327.84s &111.933s &235.21\\
\hline
\end{tabular}
\end{center}
\end{table}

\begin{figure}[t]
\begin{center}
   \includegraphics[height=4.5cm,width=6.5cm]{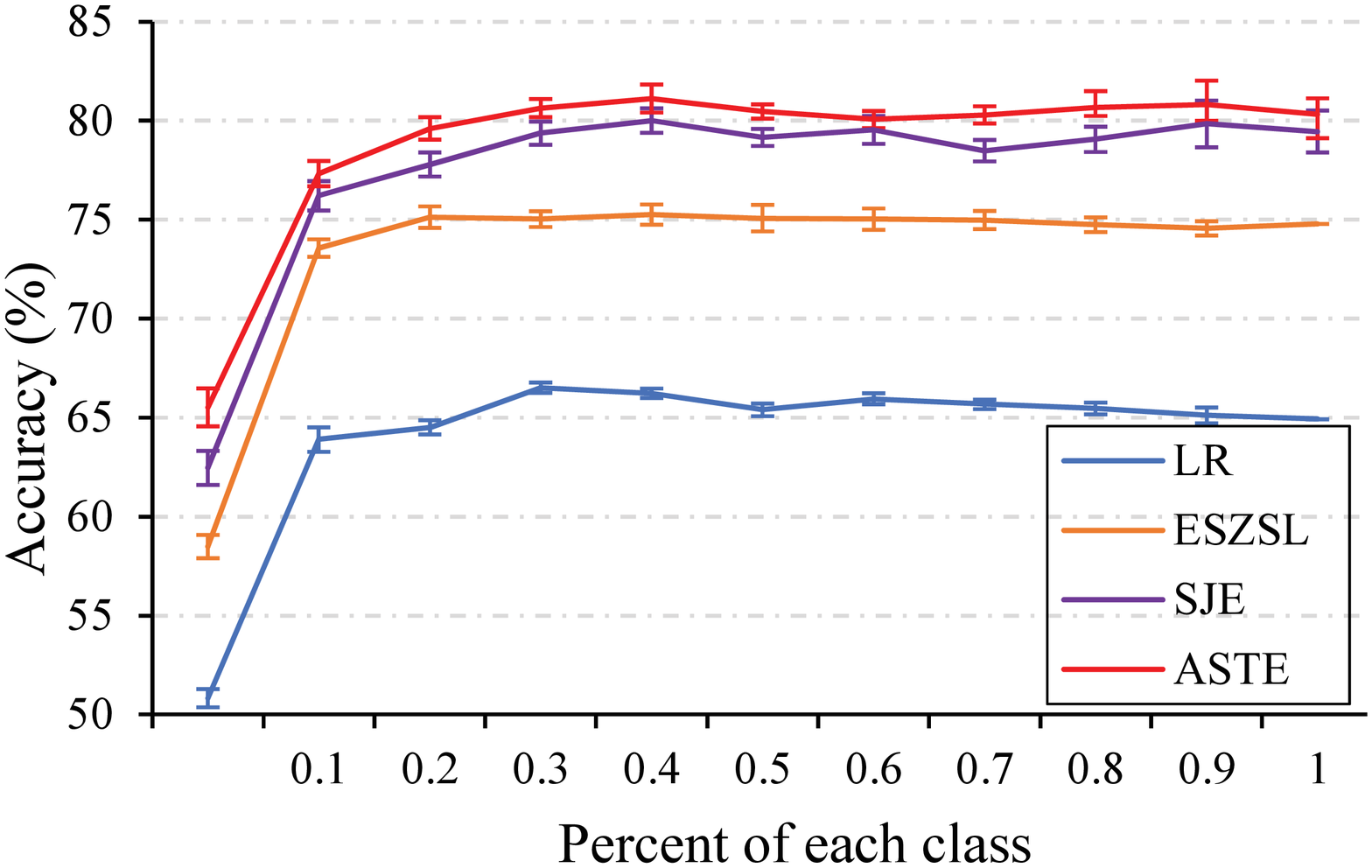}
\end{center}
   \caption{Classification performance of different approaches on AwA dataset, when varying the percent of each class number. Note: the starting point of the curve is the performance of randomly select trial.}
\label{fig:long}
\label{fig:onecol}
\end{figure}

\section{Conclusion}

In this paper, we mainly addressed the visual-semantic embedding and domain adaptation in ZSL. For the first one, we proposed an adaptive structural embedding (ASTE) method by formulating the visual-semantic embedding in an adaptively latent structural SVM framework where the reliability and discriminability of the training instances are exploited. For the second one, we presented a self-paced selective strategy (SPASS) to iteratively select the unseen instances from reliable to less reliable to gradually transfer the knowledge from
the seen domain to the unseen domain. Then, we combined ASTE and SPASS to develop a transductive ZSL approach named TASTE to progressively reinforce the discriminant capacity. Extensive experiments on three benchmark datasets have verified the superiorities of these proposed methods. Specifically, ASTE achieved the best performance on AwA dataset, and TASTE performs the best on AwA and aPY datasets. Besides, we also present a simple but effective fast training (FT) strategy to speed up the training speed of ZSL by employing the visual pattern of each class as input training data. The speed up rations are about 4 for the methods with closed-form solutions and hundreds for those without closed-form solutions.



%


\ifCLASSOPTIONcaptionsoff
  \newpage
\fi

\end{document}